\documentclass[journal]{IEEEtran}

\usepackage[inline]{enumitem}

\usepackage{url}
\def\UrlBreaks{\do\/\do-}
\usepackage{breakurl}
\usepackage{comment}
\usepackage{cite}
\usepackage{graphicx}

\usepackage{hyperref}
\usepackage[capitalize,nameinlink]{cleveref}
\usepackage{xcolor}
\hypersetup{
	unicode=true,
        bookmarksnumbered=true,
	colorlinks=true,
	linkcolor={red!70!black},
	citecolor={red!70!black},
	urlcolor={blue!70!black},
	pdfborder={0 0 0}
}
\usepackage{flushend}

\begin{document}

\title{Towards Greener LLMs: Bringing Energy-Efficiency to the Forefront of LLM Inference}

\author{
Jovan Stojkovic, Esha Choukse\textsuperscript{\textdagger}, Chaojie Zhang\textsuperscript{\textdagger}, Inigo Goiri\textsuperscript{\textdagger}, Josep Torrellas\\
    \textit{University of Illinois at Urbana-Champaign} \qquad \textsuperscript{\textdagger}\textit{Microsoft Azure Research - Systems}\\
}

\maketitle

\begin{abstract}
With the ubiquitous use of modern large language models (LLMs) across industries, the inference serving for these models is ever expanding.
Given the high compute and memory requirements of modern LLMs, more and more top-of-the-line GPUs are being deployed to serve these models.
Energy availability has come to the forefront as the biggest challenge for data center expansion to serve these models.
In this paper, we present the trade-offs brought up by making energy efficiency the primary goal of LLM serving under performance SLOs.
We show that depending on the inputs, the model, and the service-level agreements, there are several knobs available to the LLM inference provider to use for being energy efficient.
We characterize the impact of these knobs on the latency, throughput, as well as the energy.
By exploring these trade-offs, we offer valuable insights into optimizing energy usage without compromising on performance, thereby paving the way for sustainable and cost-effective LLM deployment in data center environments.

\end{abstract}


\section{Introduction}
\IEEEPARstart{M}{odern} generative large language models (LLMs) are turning ubiquitous in their use-cases, leading to large-scale inference deployments. This has lead the datacenter expansion to hit an energy wall for the foreseeable future~\cite{energywall}, further delaying the green energy promises. 
At the same time, such large scale deployments of these models present a unique opportunity to optimize the service for energy efficiency with huge impacts.
Previous work in the field of LLM inference platforms has focused on improving latency and throughput of the serving platforms.
However, we note that just like any other service, even LLM inference has periods of lower utilization, leading to slack time compared to the latency and throughput SLOs. 
We use this insight to explore various energy efficiency knobs available to the inference service provider.

LLM inference environments have various sources of inefficiency.
Prior work attacked some of the largest ones, such as 
inefficient request scheduling and batching~\cite{sarathi,splitwise,orca}, 
memory management and key-value caching of intermediate results~\cite{pagedattention,alizadeh2024llm}, speculative decoding~\cite{miao2024specinfer} or model parallelism~\cite{alpaserve}.

However, one aspect that has been largely overlooked is the energy consumption of LLM inference servers.
Despite the widespread use of LLMs and their serving engines, there is a notable absence of a comprehensive framework for managing energy in these systems.
While existing research has highlighted the unique performance challenges of LLM inference servers, 
understanding how these issues translate into power and energy consumption, as well as designing effective power management strategies, remains largely unexplored. 
%This gap is particularly concerning given the increasing reliance on LLM inference servers across various applications, contributing significantly to data center loads and consequently to global energy consumption and carbon emissions.  
Advancing research in this area is critical,
as LLM services are an increasing fraction of data center loads~\cite{money1},
and data centers contribute substantially to the world energy
consumption~\cite{energyCostDataCenters,globalElect} and  
carbon footprint~\cite{act,carbon}.

To address this shortcoming with current LLMs and inference platforms, this paper performs a thorough characterization of energy consumption in LLM inference environments under various settings. 
The goal of the characterization is to generate datasets that can provide insights and guide the design of future energy management frameworks specifically designed for LLM inference servers.

Our characterization shows that LLM inference environments pose a set of  challenges   not met
by the existing power and energy management schemes designed for traditional latency-critical data-center 
applications (~\cite{retail,gemini,rubik,pegasus,eetl,adrenaline,twig,parties}).
First, 
a fundamental challenge lies in the variable nature of loads encountered by LLM inference servers, akin to user-facing applications.
However, the requests for LLM models exhibit a remarkable diversity, with inputs ranging from short (a few tokens) to long (a few thousands of tokens)
and outputs displaying similar variability. 
Longer inputs necessitate increased GPU parallelism, resulting in extended \emph{prefill} phases, 
while longer outputs induce multiple iterations and elongated \emph{decode} phases. 
It is known that prefill phase puts more pressure on the compute resources, while decode phase puts more pressure on the memory subsystem~\cite{splitwise,polca}.
Consequently, bursts of requests of one type manifest distinct behaviors compared to those of another type, complicating load and energy management strategies.

Second,
LLM inference servers can be organized into various configurations concerning the degree and type of parallelism, batch sizes, and GPU frequencies. 
Each configuration may be optimal for specific system states. 
For instance, low loads of requests with short inputs and outputs may operate at lower GPU frequencies and with fewer GPUs (smaller degrees of tensor/pipeline parallelism).
On the other hand, high loads of requests with long inputs and outputs require high GPU frequencies and many GPUs (larger degrees of tensor/pipeline parallelism).
Yet the rapid fluctuations in load and 
software-level overheads
exacerbate the challenges associated with transitioning between configurations. For instance, adjusting GPU frequency via the CPU controller incurs significant stalls, while re-sharding the model into a new pipeline or tensor organization proves to be prohibitively expensive in terms of computational resources and time.

To effectively manage energy consumption in LLM inference environments, it is imperative to develop strategies that accommodate the dynamic and heterogeneous nature of workload characteristics. These strategies must incorporate mechanisms for adaptive resource allocation, ensuring that computational resources are efficiently utilized in response to evolving workload demands. Additionally, optimizations aimed at minimizing the overhead of configuration changes, such as GPU frequency adjustments and model reorganization, are vital for enhancing energy efficiency without compromising inference performance. By addressing these challenges, we can pave the way for the development of sustainable and energy-efficient LLM inference systems, thereby facilitating their widespread adoption across various domains.

Contributions of this paper are as follows:
\begin{itemize}
    \item Characterization of the LLM inference environments from the perspective of energy efficiency.
    \item Analysis of available knobs to LLM inference servers and their impact on the performace-energy trade-off.
    \item An outline of the requirements for an energy-efficient LLM inference framework.
\end{itemize}

\section{Background}

\subsection{LLM overview}

Modern LLMs are predominantly built upon transformer architectures~\cite{transformer}, which have revolutionized natural language processing tasks. 
These transformer models leverage attention mechanisms and multi-layer-perceptron (MLP) layers to effectively 
process inputs and generate corresponding outputs. 
The architecture of transformer-based LLMs can vary, with configurations such as encoder-only~\cite{encoderOnly}, 
decoder-only~\cite{decoderOnly}, 
or encoder-decoder~\cite{encodeDecode} models. 
In encoder-only models, the input text is processed to create contextualized representations, which are then utilized for downstream tasks. 
Conversely, decoder-only models focus on generating output sequences based on given inputs, leveraging the context encoded in the input embeddings. 
Encoder-decoder models, on the other hand, combine both encoder and decoder components, enabling tasks like machine translation and text summarization where the model processes input text and generates corresponding output text. 
In this paper we focus on 
generative LLMs which are usually
either decoder-only or encoder-decoder models. 
The models are auto-regressive, where each output token is generated sequentially with a forward pass of the model.
% Within the realm of generative LLMs, which are the primary focus of this paper, decoder-only and encoder-decoder architectures are commonly employed. 

\subsection{Batching and parallelism}
LLM inference batching refers to the process of grouping multiple input sequences together and processing them simultaneously during inference, exploiting this parallelism to improve efficiency. Moreover, batching enables better hardware utilization, leveraging the capabilities of modern computational resources such as GPUs and TPUs more effectively. By enhancing inference speed and resource utilization, batching plays a crucial role in scaling up LLM deployment for various applications, and making it energy-efficient.

Another widely used mechanism to increase throughput of LLM inference is the parallelism or sharding of the model across GPUs.
Tensor parallelism divides the model's parameters across multiple devices, such as GPUs or TPUs, for parallel computation per layer in the model inference. This approach optimizes hardware utilization, accelerating inference by distributing computations in each layer. Pipeline parallelism splits the model's layers or modules into stages, executing them sequentially across different devices. By overlapping computation and communication, it minimizes idle time and maximizes throughput. Together, tensor and pipeline parallelism enable efficient and scalable LLM inference, handling massive textual data effectively.

\subsection{SLOs in LLMs today}
Modern LLMs are assessed based on performance SLOs, including Time to First Token (TTFT), Time Between Tokens (TBT), and throughput. TTFT measures the time for the model to generate the first token of the output sequence, especially important for interactive and streaming responses. TBT quantifies the time spent between each output token, as it is generated in an auto-regressive manner.
Meeting these SLOs is crucial for ensuring timely and efficient responses across a range of applications, from chatbots to translation services, demanding careful optimization of model and hardware configurations.

\subsection{Energy efficiency knobs in modern GPUs}
Modern GPUs do not offer the vast majority of efficiency knobs that the modern CPUs offer~\cite{gpuknobscal}. For instance, voltage scaling, fine-grained power gating, efficient modes, and fine-grained frequency scaling are not offered by the GPUs today.
However, they do offer frequency control at a GPU-wide granularity.
Lowering the frequency during periods of low activity reduces power consumption without sacrificing a lot of performance.
\section{Methodology}
We present detailed characterization results on a recent open-source LLM Llama-2 with 70 billion parameters~\cite{llama2}.
Previous work has shown that other models like BLOOM-176B~\cite{scao2022bloom} closely correlate with each other in performance trends~\cite{splitwise}.
We run our experiments on an NVIDIA DGX-H100~\cite{dgxh100} using vLLM~\cite{vllm}, a state-of-the-art open-source LLM inference platform.

The maximum frequency for NVIDIA H100 is 1980 MHz. We run our experiments on H100 with frequency varying between 800 MHz to 1980 MHz in jumps of 200 MHz. 
Unless specified otherwise, the experiment runs using 8-way tensor parallelism.
For latency SLOs, we choose $5\times$ the TTFT and TBT achieved when running the request alone, without any batching or queuing delay. 

\section{Energy efficiency trade-offs}
At the platform level, we explore 3 levers:
\begin{enumerate*}
    \item workload type, 
    \item batching, and 
    \item model parallelism.
\end{enumerate*}

We define the workload type as a combination of the input and output lengths of the queries, combined with the total requests per second being sent to a model instance. We consider this a lever for the platform, since several scheduling and scaling algorithms at the service's cluster can define these properties as seen at a model instance.
We combine these levers with the only energy-efficiency knob available to us in modern GPUs at a node-level: frequency scaling.

\begin{figure}[t!]
\centering
\includegraphics[width=\columnwidth]{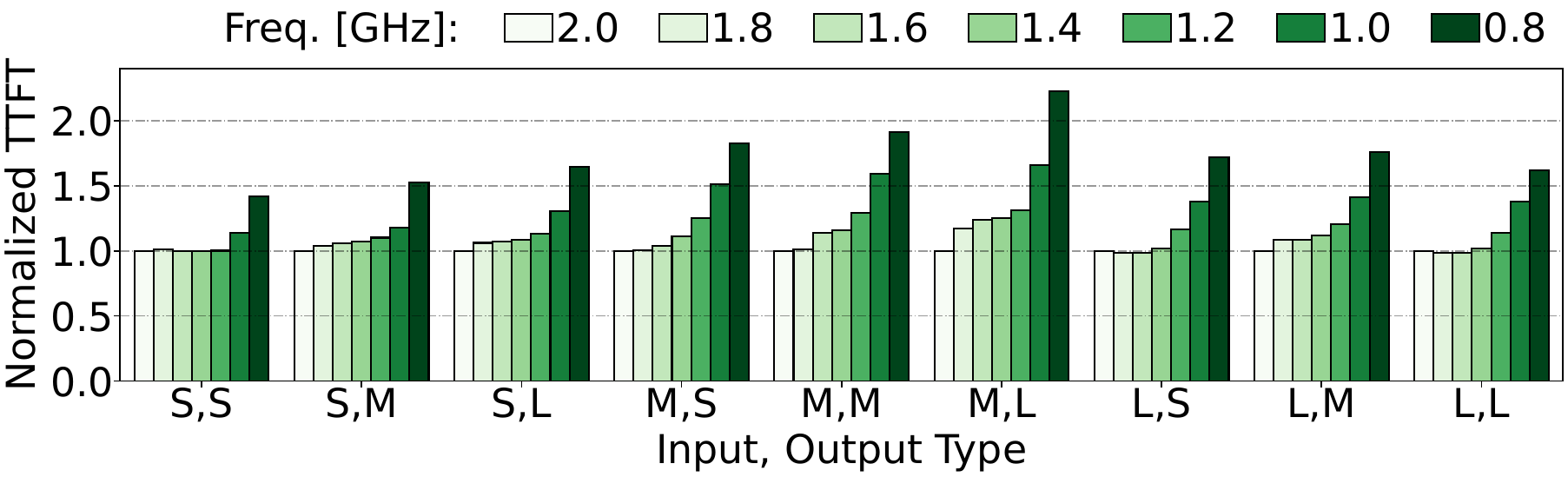}
\vspace{-6mm}
\caption{Normalized TTFT varying GPU frequencies for different inputs/outputs.}
\label{fig:ttft_freq_workloadtype}
% \vspace{-3mm}
\end{figure}

\begin{figure}[t!]
\centering
\includegraphics[width=\columnwidth]{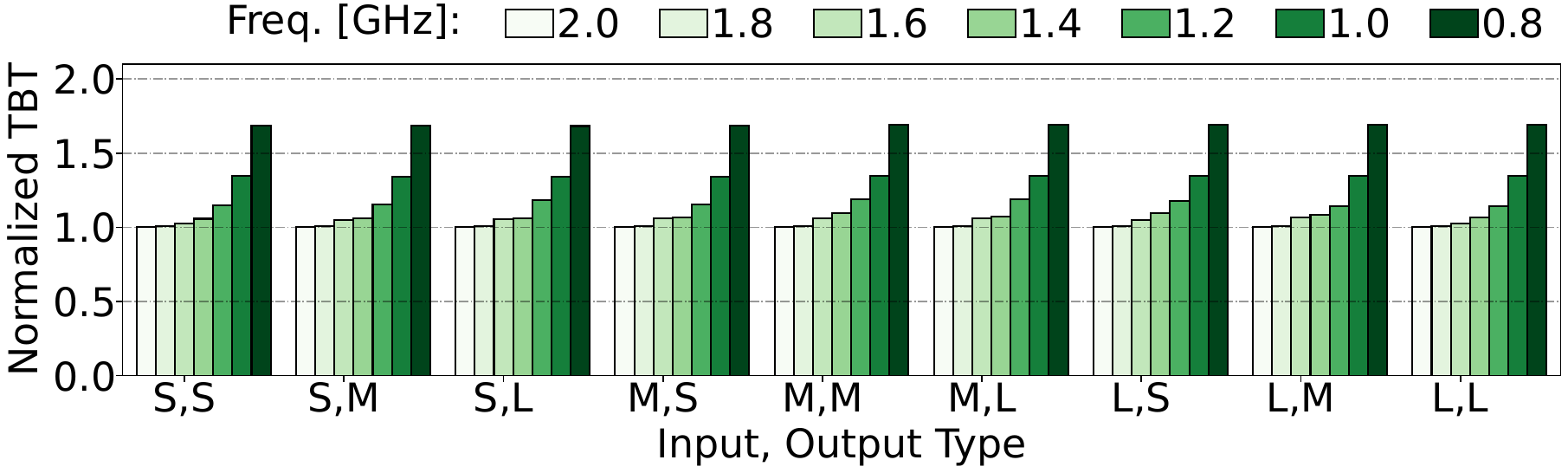}
\vspace{-6mm}
\caption{Normalized TBT varying GPU frequencies for different inputs/outputs.}
\label{fig:tbt_freq_workloadtype}
% \vspace{-3mm}
\end{figure}

\begin{figure}[t!]
\centering
\includegraphics[width=\columnwidth]{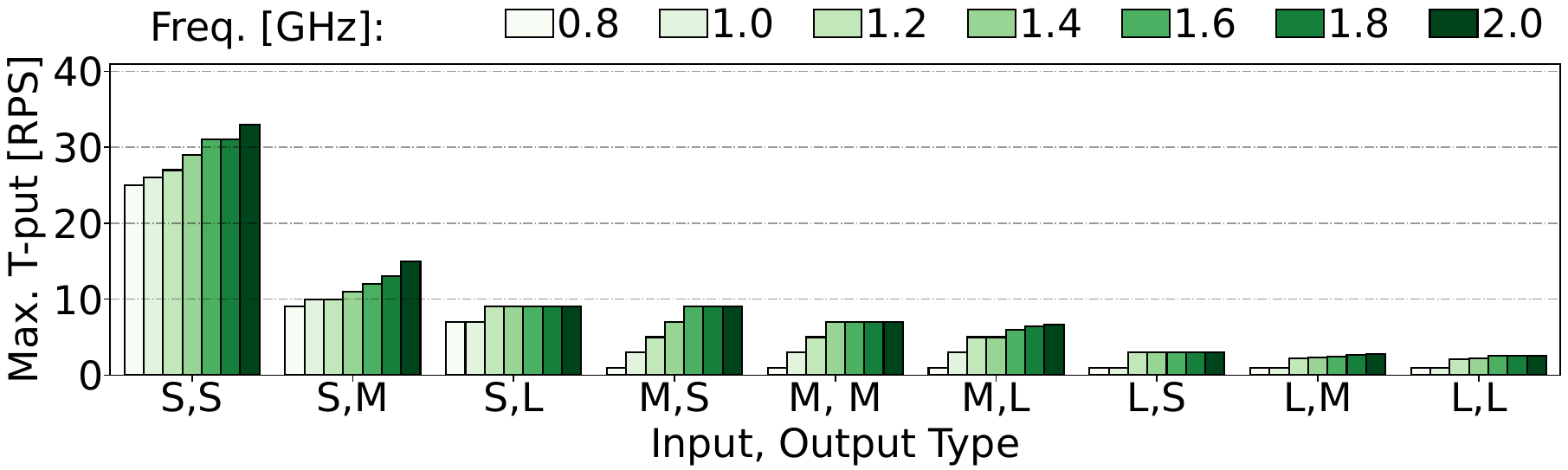}
\vspace{-6mm}
\caption{Maximum throughput of an 8-way tensor-parallel GPU LLama2 instance with different GPU frequencies for different input/output types.}
\label{fig:throughput_freq_workloadtype}
% \vspace{-3mm}
\end{figure}

\begin{figure}[t!]
\centering
\includegraphics[width=\columnwidth]{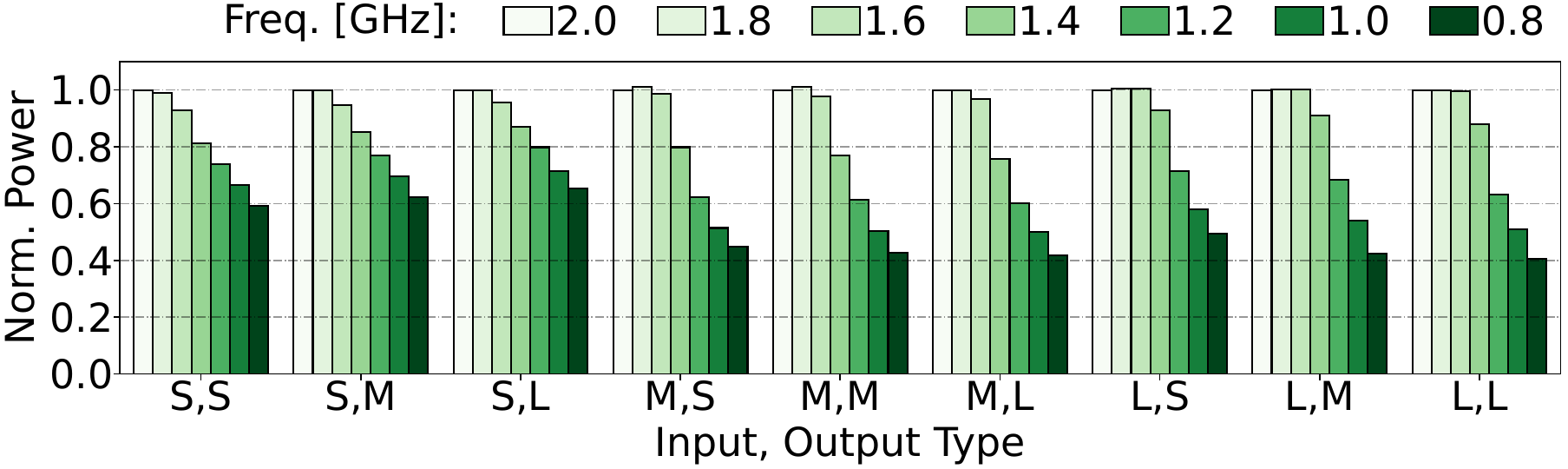}
\vspace{-6mm}
\caption{Normalized power consumption of an 8-way tensor-parallel GPU LLama2 instance with different GPU frequencies for different request types.}
\label{fig:power_freq_workloadtype}
% \vspace{-3mm}
\end{figure}

\begin{figure}[t!]
\centering
\includegraphics[width=\columnwidth]{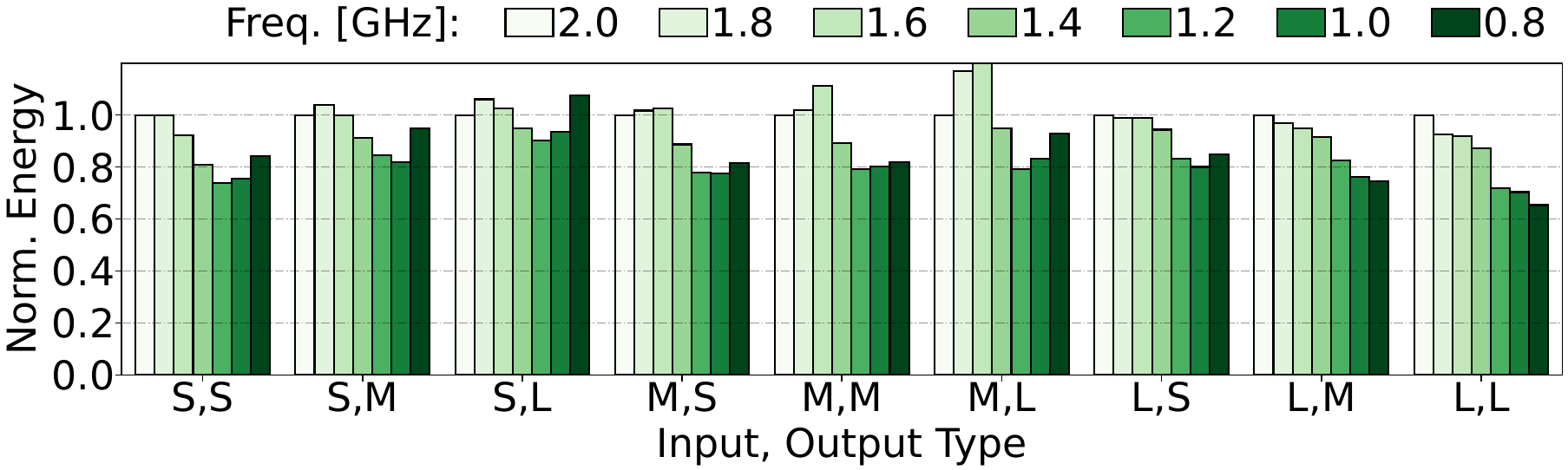}
\vspace{-6mm}
\caption{Normalized energy consumption of an 8-way tensor-parallel GPU LLama2 instance with different GPU frequencies for different request types.}
\label{fig:energy_freq_workloadtype}
% \vspace{-3mm}
\end{figure}

\subsection{Impact of workload type}
We divide the workload into buckets based on the input and output length in number of tokens. Both input and output lengths are divided into three buckets: Small (100 tokens for input and 50 tokens for output), Medium (500 and 128 tokens for input and output respectively), and Large (1024 tokens for input, 256 tokens for output).
Combining the input and output lengths from different buckets gives us 9 types of input/output workload types.
We run a streaming workload through the model instance.

%\subsubsection{Frequency response}
%\newline

\textbf{Latency}
\Cref{fig:ttft_freq_workloadtype,fig:tbt_freq_workloadtype} show the impact of the frequency setting on the TTFT and TBT latency metrics for the different workload types.
As the input length increases, the computational intensity of the prefill phase increases.
Therefore, we see a clear pattern, where the TTFT gets increasingly impacted by frequency and lowering as the prompt length increases. 
In fact, the large inputs increase the computational intensity enough to cause throttling due to power overdraw - this reduces the impact of frequency capping on large inputs.
Furthermore, longer output lengths running at lower frequency increase the queuing time, adding to the TTFT.
On the other hand, the decode phase is memory bound, and growing the input or output length has an imperceptible impact on the TBT's frequency response (\Cref{fig:tbt_freq_workloadtype}).

\textbf{Throughput}
\Cref{fig:throughput_freq_workloadtype} shows the maximum throughput achievable under the SLOs for various workload types while changing their frequency.
We note that the throughput is heavily affected by both the input and output lengths.
Longer inputs lead to higher TBT for the requests that get their decode phase batched with the prefill phase.
Longer outputs lead to queuing delay as the model instance spends more number of iterations on each request.
The small input and small output setup achieves the highest throughput and reducing the frequency by half, only reduces the throughput by $\sim$20\%.

%\subsubsection{Designing for energy}
\textbf{Energy}
\Cref{fig:power_freq_workloadtype} shows the maximum power draw for different workload types with frequency capping.
Each data point is shown at a medium load the corresponding configuration can support.
Comparing \Cref{fig:power_freq_workloadtype} with ~\Cref{fig:ttft_freq_workloadtype,fig:tbt_freq_workloadtype,fig:throughput_freq_workloadtype} shows that we can achieve $\sim$20\% lower power for most configurations without any impact to the latency or throughput.
Furthermore, if the workload is going through a low utilization phase, the power required can be reduced even further, at no impact to the workload.

\Cref{fig:energy_freq_workloadtype} shows the corresponding energy consumption.
It is evident that optimizing for power vs energy vs performance would lead to very different frequency configurations for the inference platform.

\subsection{Impact of parallelism}
Next, we vary the tensor parallelism degree across different number of GPUs under medium load and medium input/output workload types. Tensor parallelism divides the KV heads of the model equally across the GPUs.
Therefore, we use tensor parallelism degrees of 2, 4, and 8 within a single DGX-H100 node, and name these configurations TP2, TP4, and TP8.

%\subsubsection{Frequency response} 

\begin{figure}[t!]
\centering
\includegraphics[width=\columnwidth]{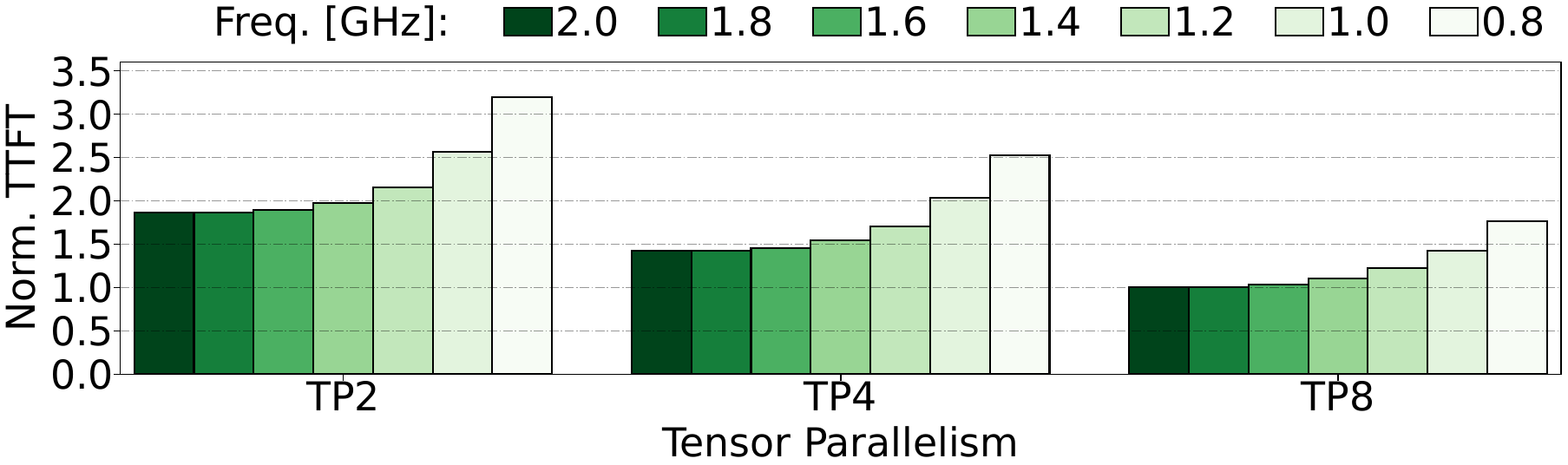}
\vspace{-6mm}
\caption{TTFT of an GPU LLama2 instance under medium load and medium input/output request types with different GPU frequencies for different tensor parallelism strategies.}
\label{fig:ttft_freq_tp}
% \vspace{-3mm}
\end{figure}

\begin{figure}[t!]
\centering
\includegraphics[width=\columnwidth]{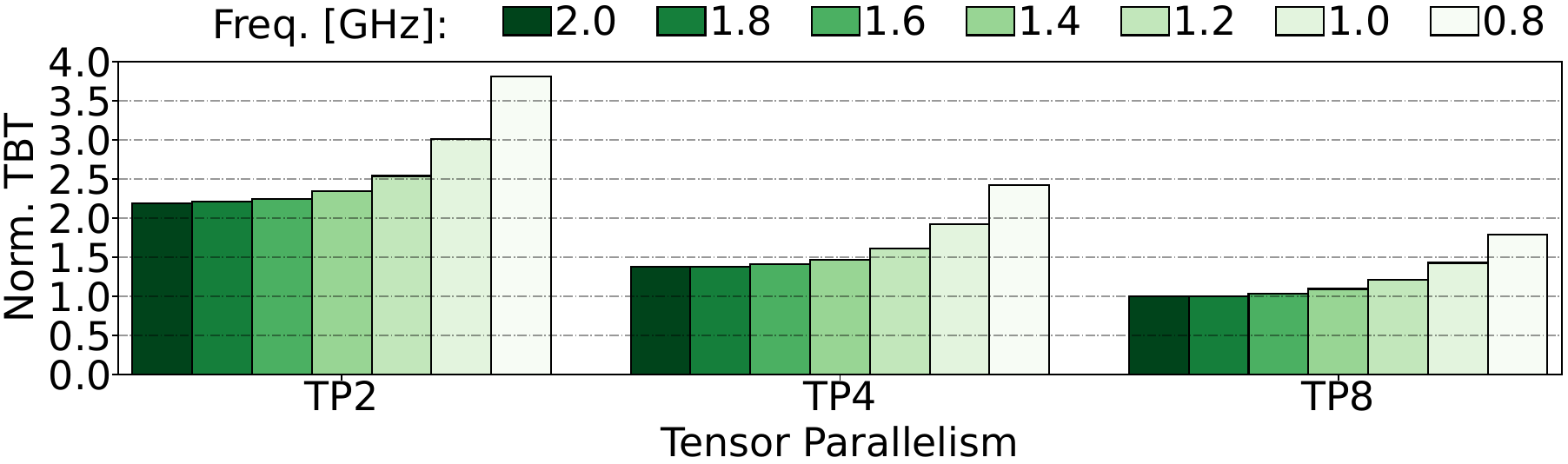}
\vspace{-6mm}
\caption{TBT of a GPU LLama2 instance under medium load and medium input/output request types with different GPU frequencies for different tensor parallelism strategies.}
\label{fig:tbt_freq_tp}
% \vspace{-3mm}
\end{figure}

\textbf{Latency}
\Cref{fig:ttft_freq_tp,fig:tbt_freq_tp} show that increasing parallelism reduces both TTFT and TBT as tensor parallelism effectively parallelizes computation within the layers.
However, as communication overhead also grows, larger parallelism does not achieve linear latency reduction.
As computational intensity remains high during prefill phase, TTFT exhibits similar frequency responses across parallelisms.
In contrast, increasing parallelism reduces the computation on each GPU during decode phase, TBT's frequency impact further decreases.

\begin{figure}[t!]
\centering
\includegraphics[width=\columnwidth]{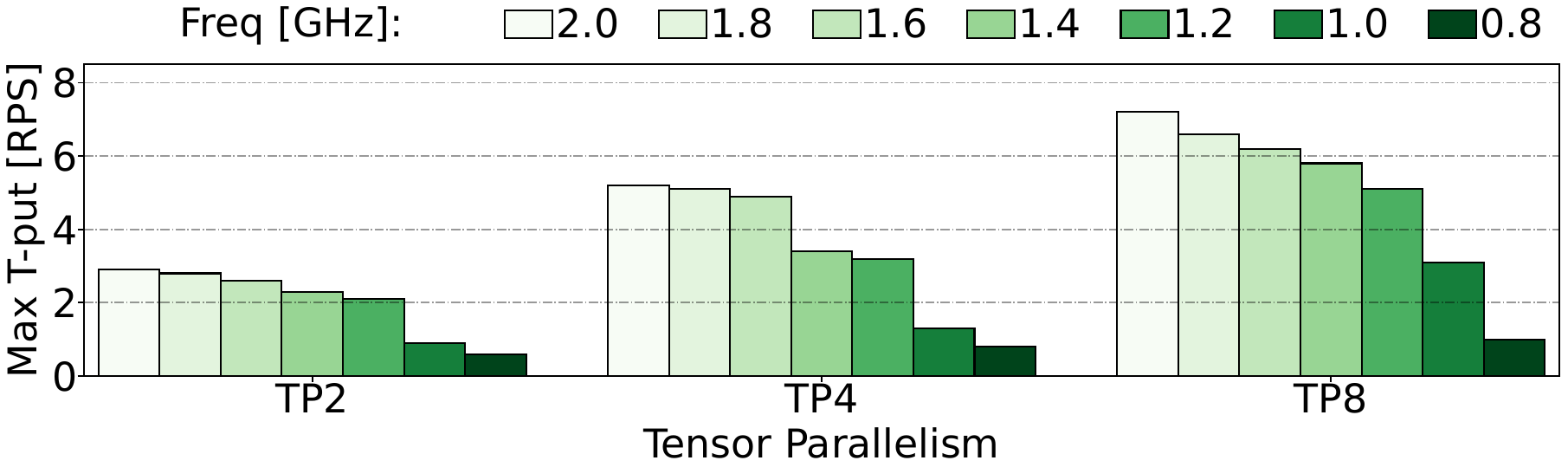}
\vspace{-6mm}
\caption{Maximum throughput of a LLama2 instance under medium inputs/outputs request types with different GPU frequencies for different tensor parallelism strategies.}
\label{fig:throughput_4tp_workloadtype}
% \vspace{-3mm}
\end{figure}

\textbf{Throughput}
\Cref{fig:throughput_4tp_workloadtype} shows the maximum throughput achievable by each of the parallelism configurations under latency SLOs.
As expected, the increase in throughput going from TP2 to TP4 is much higher (75\%) compared to the increase from TP4 to TP8 (40\%).
In either case, if optimizing for cost per request, as long as TP2 meets the latency SLOs, having 2 instances of TP2 is better than TP4. Similarly, two instances of TP4 would be better than TP8.

\begin{figure}[t!]
\centering
\includegraphics[width=\columnwidth]{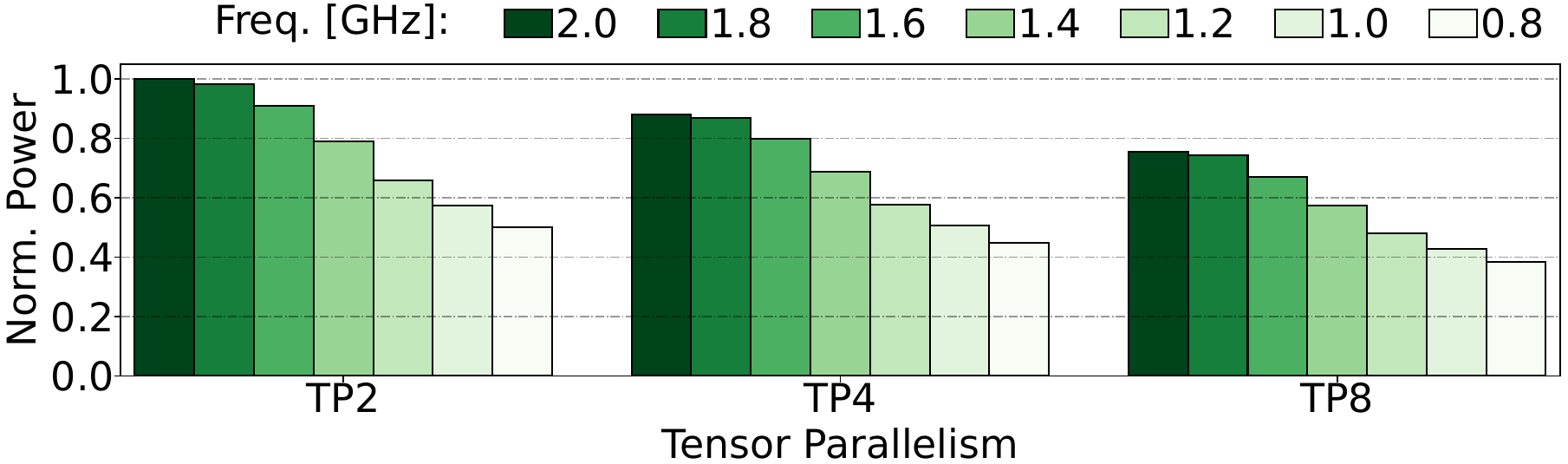}
\vspace{-6mm}
\caption{Normalized per-GPU power consumption of a LLama2 instance under medium load and medium input/output request types with different GPU frequencies for different levels of tensor-parallelism.}
\label{fig:power_freq_tp}
% \vspace{-3mm}
\end{figure}

\begin{figure}[t!]
\centering
\includegraphics[width=\columnwidth]{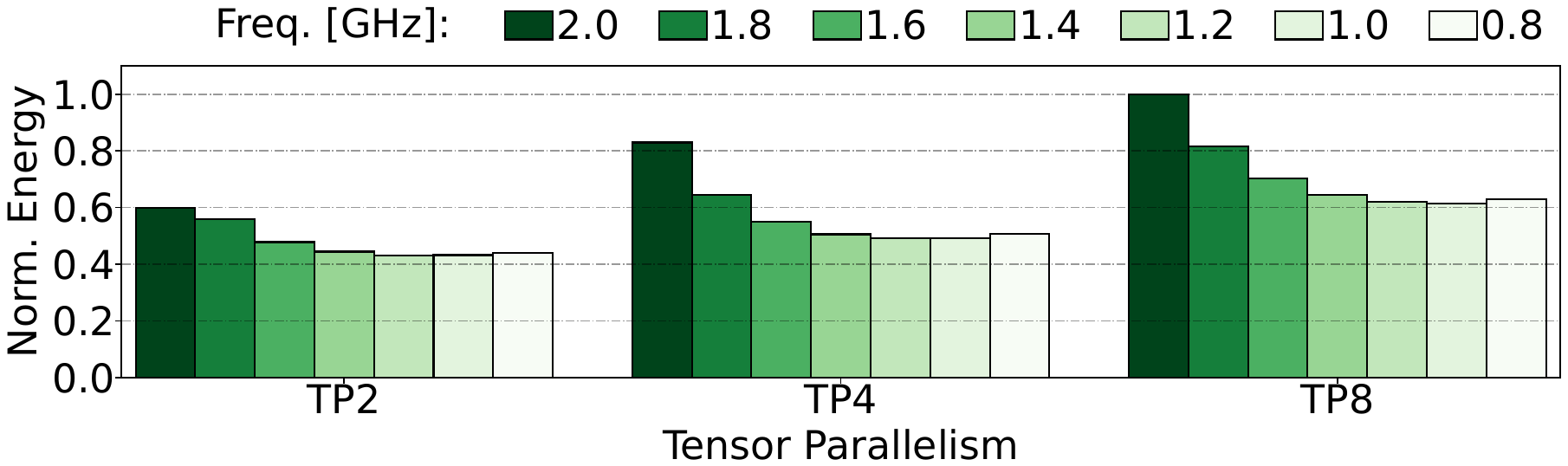}
\vspace{-6mm}
\caption{Normalized total energy consumption of a LLama2 instance under medium load and medium input/output request types with different GPU frequencies for different levels of tensor-parallelism.}
\label{fig:energy_freq_tp}
% \vspace{-3mm}
\end{figure}

%\subsubsection{Designing for energy}
\textbf{Energy}
\Cref{fig:power_freq_tp,fig:energy_freq_tp} show the maximum power and total energy consumption at medium load for each configuration. These are particularly interesting, since the normalized total energy for TP2 is only 40\% lower than TP8, which being able to serve only 60\% fewer requests. 
Furthermore, most cloud environments today only allow full node access (8 GPUs) to DGX-H100 nodes.
This means that in times of lower throughput needs, it is more energy-efficient to run TP8 than TP2!

\subsection{Impact of batching}
As mentioned before, we use mixed batching of prefill and decode phases.
We increase the maximum allowed batch size and observe the patterns in performance and energy with frequency scaling.

%\subsubsection{Frequency response}

\begin{figure}[t!]
\centering
\includegraphics[width=\columnwidth]{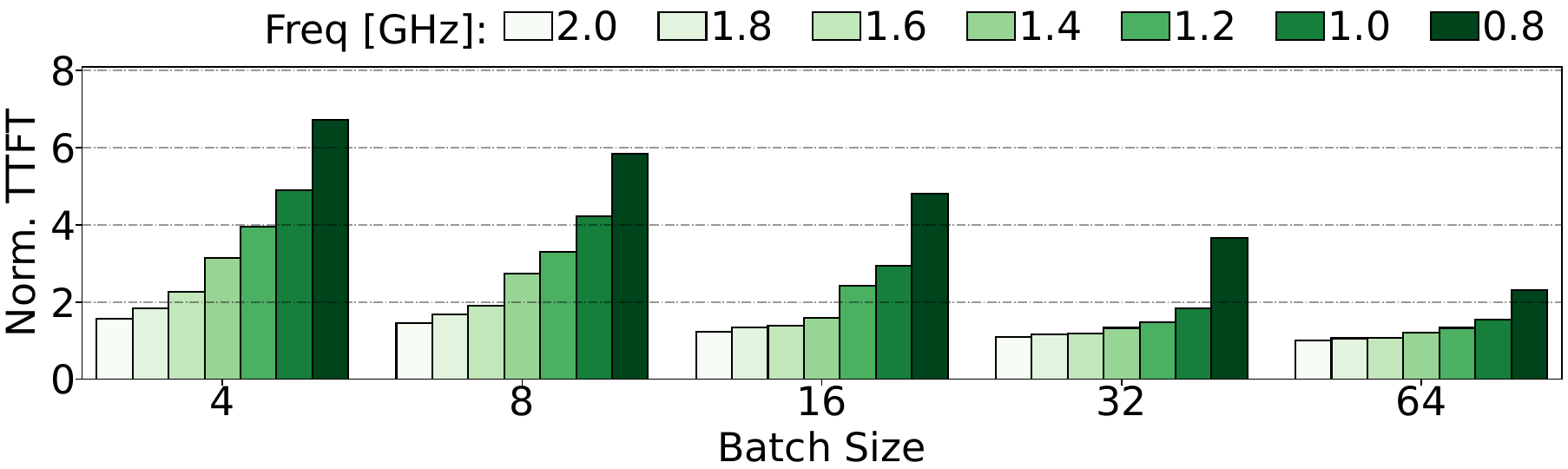}
\vspace{-6mm}
\caption{Normalized TTFT of an 8-way tensor-parallel LLama2 instance under medium load and medium inputs/outputs request types with different GPU frequencies for different batch sizes.}
\label{fig:ttft_batch}
% \vspace{-3mm}
\end{figure}

\begin{figure}[t!]
\centering
\includegraphics[width=\columnwidth]{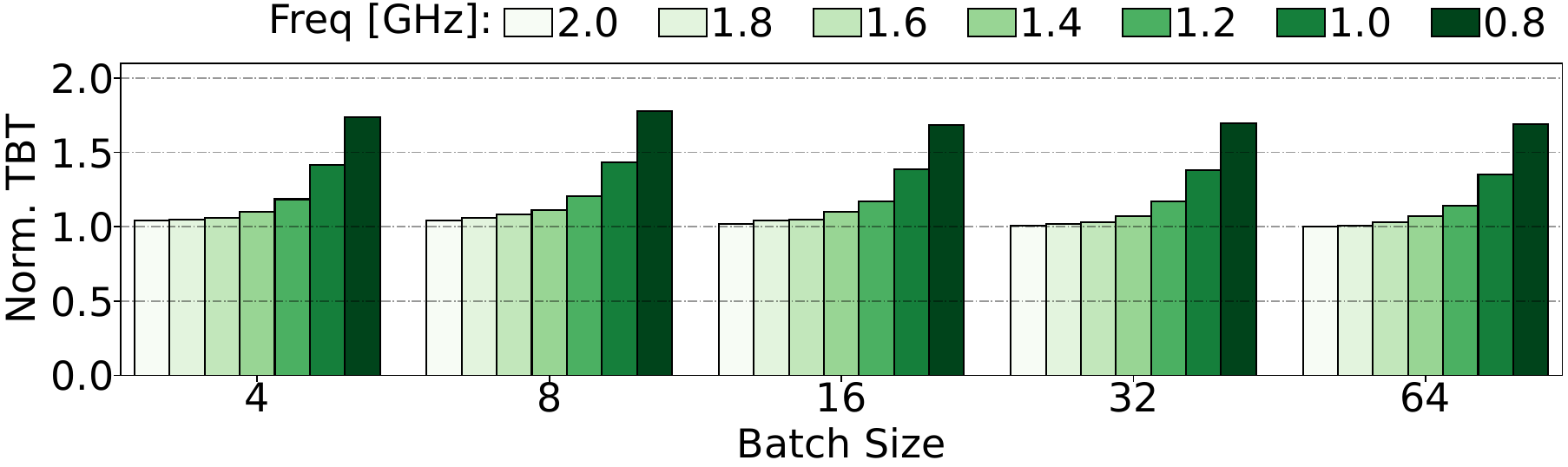}
\vspace{-6mm}
\caption{Normalized TBT of an 8-way tensor-parallel LLama2 instance under medium load and medium inputs/outputs request types with different GPU frequencies for different batch sizes.}
\label{fig:tbt_batch}
% \vspace{-3mm}
\end{figure}

\textbf{Latency}
\Cref{fig:ttft_batch,fig:tbt_batch} show the TTFT and TBT when batching.
At maximum frequency, the TTFT slightly decreases as the maximum batch size is increased. This is due to reduction in queuing of the request before it gets batched in for inference. At lower frequencies, this effect is even more pronounced. With very low load, we see the opposite trend, where the TTFT increases due to increased computational complexity as the batch size increases.
Since queuing delay does not impact TBT due to no preemption in this setup, we do not see such impact on the decode phase. Additionally, decode phase being memory-bound does not experience slow down as the batch size increases up to 64.

\begin{figure}[t!]
\centering
\includegraphics[width=\columnwidth]{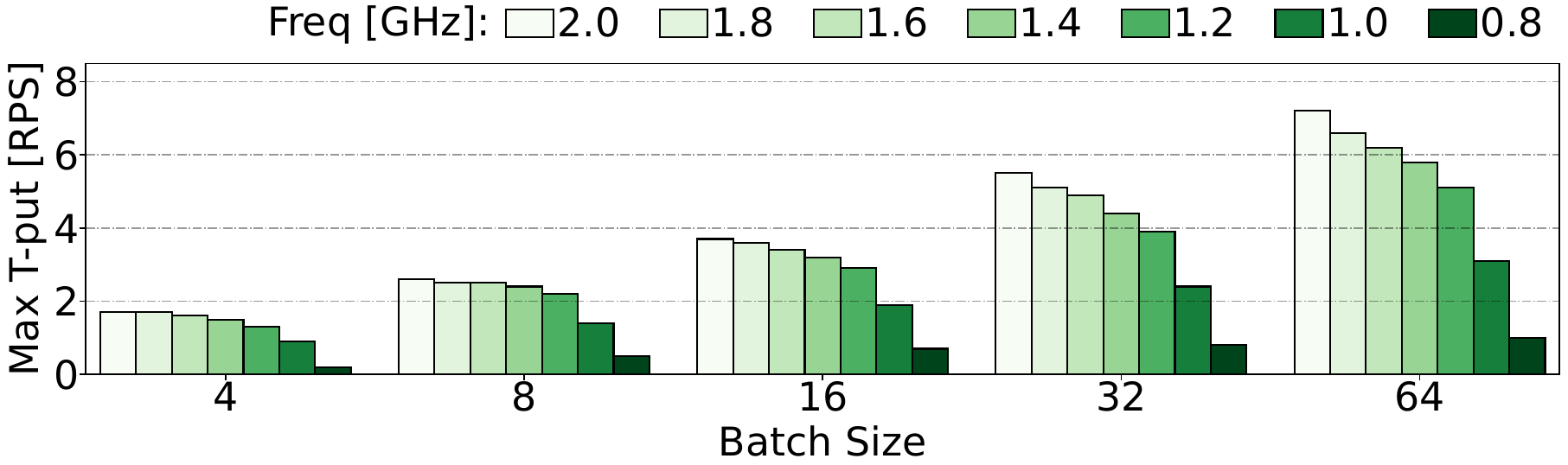}
\vspace{-6mm}
\caption{Maximum throughput of an 8-way tensor-parallel LLama2 instance under medium inputs/outputs request types with different GPU frequencies for different batch sizes.}
\label{fig:tput_batch}
% \vspace{-3mm}
\end{figure}

\textbf{Throughput}
\Cref{fig:tput_batch} shows the maximum throughput for different batch sizes.
Since larger batches can lead to TTFT SLO misses, the throughput increase on doubling the batch size is not double. In fact, under SLOs, a batch size of 64 has only 7$\times$ higher throughput than a batch size of 4.

\begin{figure}[t!]
\centering
\includegraphics[width=\columnwidth]{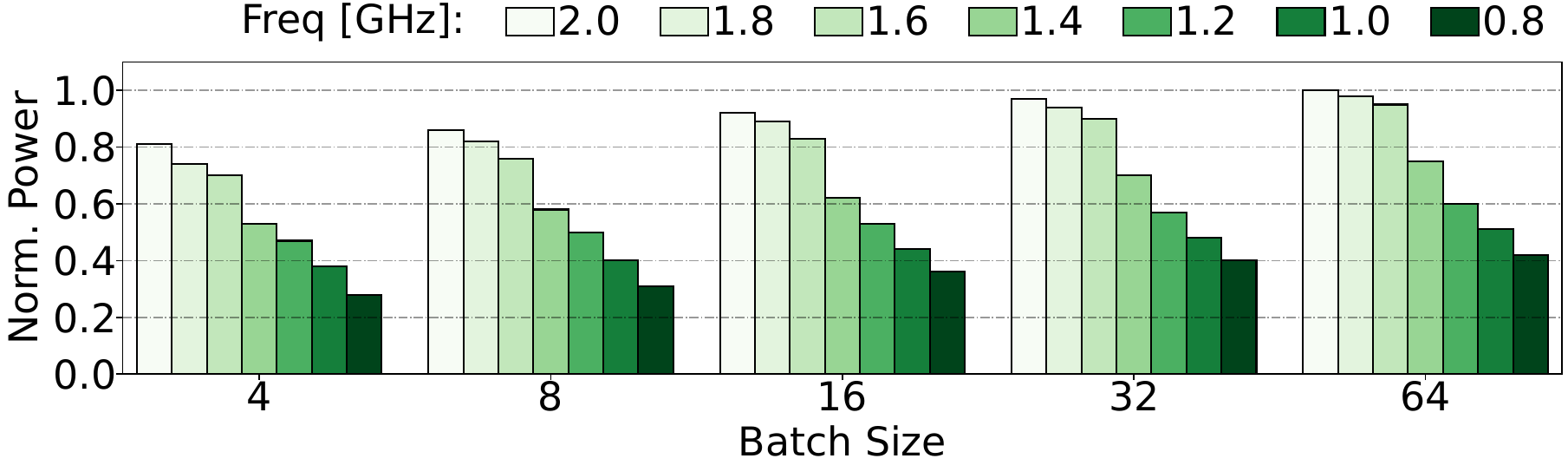}
\vspace{-6mm}
\caption{Normalized power consumption of an 8-way tensor-parallel LLama2 instance under medium load and medium inputs/outputs request types with different GPU frequencies for different batch sizes.}
\label{fig:power_batch}
% \vspace{-3mm}
\end{figure}

\begin{figure}[t!]
\centering
\includegraphics[width=\columnwidth]{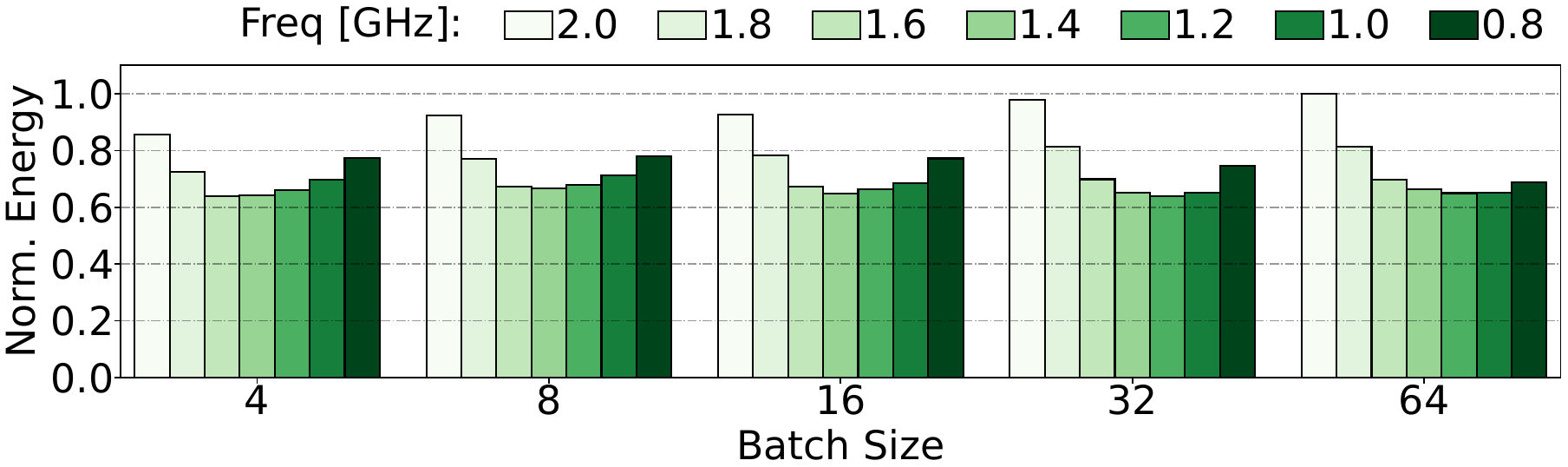}
\vspace{-6mm}
\caption{Normalized energy consumption of an 8-way tensor-parallel LLama2 instance under medium load and medium inputs/outputs request types with different GPU frequencies for different batch sizes.}
\label{fig:energy_batch}
% \vspace{-3mm}
\end{figure}

%\subsubsection{Designing for energy}
\textbf{Energy}
\Cref{fig:power_batch,fig:energy_batch} show the normalized maximum power and total energy observations. For most batch sizes, running the GPUs at $1.6 GHz$ instead of $2 GHz$ yields about the same throughput at less than 80\% of the energy.
Additionally, during phases of lower throughput needs in a service, reduce the maximum batch size can reduce consumed energy by up to 15\%.

\begin{comment}
\begin{figure}
\centering
\includegraphics[width=\columnwidth]{figures/batchPower1_crop.pdf}
\vspace{-6mm}
\caption{Per-GPU power consumption of an 8-way tensor-parallel LLama2 instance under large inputs with different batch sizes for different loads.}
\label{fig:power_batchLarge}
% \vspace{-3mm}
\end{figure}

\begin{figure}
\centering
\includegraphics[width=\columnwidth]{figures/batchTTFT1_crop.pdf}
\vspace{-6mm}
\caption{TTFT of an 8-way tensor-parallel LLama2 instance under small inputs with different batch sizes for different loads.}
\label{fig:ttft_batchSmall}
% \vspace{-3mm}
\end{figure}

\begin{figure}
\centering
\includegraphics[width=\columnwidth]{figures/batchPower_crop.pdf}
\vspace{-6mm}
\caption{Per-GPU power consumption of an 8-way tensor-parallel LLama2 instance under small inputs with different batch sizes for different loads.}
\label{fig:power_batchSmall}
% \vspace{-3mm}
\end{figure}
\end{comment}
\section{Related work}

\textbf{LLM workload characterization}
A large body of work has characterized LLM inference workloads focusing on performance and utilization~\cite{charUtil,charHW}. 
Moreover, some works have looked at LLM workloads from power\cite{polca}, carbon~\cite{llmcarbon}, and energy~\cite{wordstowatts} perspective. 
In light of this, it becomes evident that a holistic understanding of LLM inference workloads necessitates a comprehensive examination of their performance, energy efficiency, and environmental implications. 
Such an approach not only enables better resource management but also facilitates the identification of actionable optimization strategies and fine-tuning mechanisms for LLM inference serving.
By bridging the gap between performance metrics, energy efficiency considerations, and actionable optimization strategies, this paper provides insights to researchers and practitioners 
to
navigate the complexities of LLM inference workloads more effectively, ultimately contributing to more sustainable and efficient deployment of these 
models in real-world applications.

\textbf{Energy efficiency with hardware accelerators}
Other works have proposed the use of different models~\cite{phislm,snapea} and different hardware~\cite{fpga1,kim2023stack,Armeniakos_2022} for energy efficient execution of transformer-based architectures. We focus on exposing and understanding energy efficiency of ubiquitous LLM inference infrastructures and immediately actionable knobs that require no changes in server's hardware nor model's architecture. 

% include the use of different models and different hardware for energy efficiency - but we focus on the largest slice, immediately actionable

\textbf{Efficient LLM inference serving}
Many recent works propose to optimize cluster and node-level scheduling~\cite{sarathi, orca, splitwise}, memory and key-value cache management \cite{vllm, alizadeh2024llm,flashattention}, and model parallelism\cite{alpaserve} to improve inference efficiency. While these works focus on latency and throughput improvement, energy-efficient LLM serving exhibits distinct trade-offs and thus requires comprehensive understanding and solutions.
\section{Conclusion and Future Work}
In this work, we present a characterization of the impact of energy-efficiency knobs on various levers available to a modern LLM serving platform.
We offer valuable insights into the difference between performance-optimized vs energy-optimized designs on the widely adopted model and hardware.
We further show that there are platform decisions that can be made towards better energy-efficiency at no impact to the cost and performance. 
With this, we pave the way for orchestration and cluster and node-level scheduling work with more holistic optimization functions in future.

\bibliographystyle{IEEEtranS}
\bibliography{refs}

% Generated by IEEEtranS.bst, version: 1.14 (2015/08/26)
\begin{thebibliography}{10}
\providecommand{\url}[1]{#1}
\csname url@samestyle\endcsname
\providecommand{\newblock}{\relax}
\providecommand{\bibinfo}[2]{#2}
\providecommand{\BIBentrySTDinterwordspacing}{\spaceskip=0pt\relax}
\providecommand{\BIBentryALTinterwordstretchfactor}{4}
\providecommand{\BIBentryALTinterwordspacing}{\spaceskip=\fontdimen2\font plus
\BIBentryALTinterwordstretchfactor\fontdimen3\font minus \fontdimen4\font\relax}
\providecommand{\BIBforeignlanguage}[2]{{%
\expandafter\ifx\csname l@#1\endcsname\relax
\typeout{** WARNING: IEEEtranS.bst: No hyphenation pattern has been}%
\typeout{** loaded for the language `#1'. Using the pattern for}%
\typeout{** the default language instead.}%
\else
\language=\csname l@#1\endcsname
\fi
#2}}
\providecommand{\BIBdecl}{\relax}
\BIBdecl

\bibitem{sarathi}
A.~Agrawal, A.~Panwar, J.~Mohan, N.~Kwatra, B.~S. Gulavani, and R.~Ramjee, ``{SARATHI: Efficient LLM Inference by Piggybacking Decodes with Chunked Prefills},'' 2023.

\bibitem{snapea}
V.~Akhlaghi, A.~Yazdanbakhsh, K.~Samadi, R.~K. Gupta, and H.~Esmaeilzadeh, ``{SnaPEA: Predictive Early Activation for Reducing Computation in Deep Convolutional Neural Networks},'' in \emph{Proceedings of the ACM/IEEE 45th Annual International Symposium on Computer Architecture (ISCA)}, 2018.

\bibitem{alizadeh2024llm}
K.~Alizadeh, I.~Mirzadeh, D.~Belenko, K.~Khatamifard, M.~Cho, C.~C.~D. Mundo, M.~Rastegari, and M.~Farajtabar, ``{LLM in a flash: Efficient Large Language Model Inference with Limited Memory},'' 2024.

\bibitem{globalElect}
A.~Andrae and T.~Edler, ``{On Global Electricity Usage of Communication Technology: Trends to 2030},'' \emph{Challenges}, vol.~6, 2015.

\bibitem{Armeniakos_2022}
\BIBentryALTinterwordspacing
G.~Armeniakos, G.~Zervakis, D.~Soudris, and J.~Henkel, ``{Hardware Approximate Techniques for Deep Neural Network Accelerators: A Survey},'' \emph{ACM Computing Surveys}, vol.~55, no.~4, p. 1–36, Nov. 2022. [Online]. Available: \url{http://dx.doi.org/10.1145/3527156}
\BIBentrySTDinterwordspacing

\bibitem{energywall}
K.~Blunt and J.~Hiller, ``{Big Tech’s Latest Obsession Is Finding Enough Energy},'' \url{https://www.wsj.com/business/energy-oil/big-techs-latest-obsession-is-finding-enough-energy-f00055b2}.

\bibitem{parties}
S.~Chen, C.~Delimitrou, and J.~F. Mart\'{\i}nez, ``{PARTIES: QoS-Aware Resource Partitioning for Multiple Interactive Services},'' in \emph{ASPLOS}, 2019.

\bibitem{retail}
S.~Chen, A.~Jin, C.~Delimitrou, and J.~F. Martínez, ``{ReTail: Opting for Learning Simplicity to Enable QoS-Aware Power Management in the Cloud},'' in \emph{HPCA}, 2022.

\bibitem{flashattention}
T.~Dao, D.~Y. Fu, S.~Ermon, A.~Rudra, and C.~Ré, ``{FlashAttention: Fast and Memory-Efficient Exact Attention with IO-Awareness},'' 2022.

\bibitem{encoderOnly}
J.~Devlin, M.-W. Chang, K.~Lee, and K.~Toutanova, ``{BERT: Pre-training of Deep Bidirectional Transformers for Language Understanding},'' 2019.

\bibitem{llmcarbon}
A.~Faiz, S.~Kaneda, R.~Wang, R.~Osi, P.~Sharma, F.~Chen, and L.~Jiangr, ``{LLMCarbon: Modeling the end-to-end Carbon Footprint of Large Language Models},'' 2024.

\bibitem{money1}
{Forbes}, ``{Generative AI Breaks The Data Center},'' \url{https://www.forbes.com/sites/tiriasresearch/2023/05/12/generative-ai-breaks-the-data-center-data-center-infrastructure-and-operating-costs-projected-to-increase-to-over-76-billion-by-2028/?sh=5bca69067c15}.

\bibitem{phislm}
S.~Gunasekar, Y.~Zhang, J.~Aneja, C.~C.~T. Mendes, A.~D. Giorno, S.~Gopi, M.~Javaheripi, P.~Kauffmann, G.~de~Rosa, O.~Saarikivi, A.~Salim, S.~Shah, H.~S. Behl, X.~Wang, S.~Bubeck, R.~Eldan, A.~T. Kalai, Y.~T. Lee, and Y.~Li, ``Textbooks are all you need,'' 2023.

\bibitem{carbon}
U.~Gupta, Y.~Kim, S.~Lee, J.~Tse, H.~S. Lee, G.~Wei, D.~Brooks, and C.~Wu, ``{Chasing Carbon: The Elusive Environmental Footprint of Computing},'' in \emph{HPCA '21}, 2021.

\bibitem{act}
U.~Gupta, M.~Elgamal, G.~Hills, G.-Y. Wei, H.-H.~S. Lee, D.~Brooks, and C.-J. Wu, ``{ACT: designing sustainable computer systems with an architectural carbon modeling tool},'' in \emph{ISCA}, 2022.

\bibitem{eetl}
M.~E. Haque, Y.~He, S.~Elnikety, T.~D. Nguyen, R.~Bianchini, and K.~S. McKinley, ``{Exploiting Heterogeneity for Tail Latency and Energy Efficiency},'' in \emph{MICRO}, 2017.

\bibitem{adrenaline}
C.-H. Hsu, Y.~Zhang, M.~A. Laurenzano, D.~Meisner, T.~Wenisch, J.~Mars, L.~Tang, and R.~G. Dreslinski, ``{Adrenaline: Pinpointing and reining in tail queries with quick voltage boosting},'' in \emph{HPCA}, 2015.

\bibitem{charUtil}
Q.~Hu, Z.~Ye, Z.~Wang, G.~Wang, M.~Zhang, Q.~Chen, P.~Sun, D.~Lin, X.~Wang, Y.~Luo, Y.~Wen, and T.~Zhang, ``{Characterization of Large Language Model Development in the Datacenter},'' 2024.

\bibitem{rubik}
H.~Kasture, D.~B. Bartolini, N.~Beckmann, and D.~Sanchez, ``{Rubik: Fast analytical power management for latency-critical systems},'' in \emph{MICRO}, 2015.

\bibitem{fpga1}
H.~Khan, A.~Khan, Z.~Khan, L.~B. Huang, K.~Wang, and L.~He, ``{NPE: An FPGA-based Overlay Processor for Natural Language Processing},'' in \emph{Proceedings of the 2021 ACM/SIGDA International Symposium on Field-Programmable Gate Arrays}, ser. FPGA '21, 2021.

\bibitem{kim2023stack}
S.~Kim, C.~Hooper, T.~Wattanawong, M.~Kang, R.~Yan, H.~Genc, G.~Dinh, Q.~Huang, K.~Keutzer, M.~W. Mahoney, Y.~S. Shao, and A.~Gholami, ``{Full Stack Optimization of Transformer Inference: a Survey},'' 2023.

\bibitem{pagedattention}
W.~Kwon, Z.~Li, S.~Zhuang, Y.~Sheng, L.~Zheng, C.~H. Yu, J.~Gonzalez, H.~Zhang, and I.~Stoica, ``{Efficient Memory Management for Large Language Model Serving with PagedAttention},'' in \emph{SOSP}, 2023.

\bibitem{vllm}
W.~Kwon, Z.~Li, S.~Zhuang, Y.~Sheng, L.~Zheng, C.~H. Yu, J.~E. Gonzalez, H.~Zhang, and I.~Stoica, ``{Efficient Memory Management for Large Language Model Serving with PagedAttention},'' in \emph{SOSP}, 2023.

\bibitem{alpaserve}
Z.~Li, L.~Zheng, Y.~Zhong, V.~Liu, Y.~Sheng, X.~Jin, Y.~Huang, Z.~Chen, H.~Zhang, J.~E. Gonzalez, and I.~Stoica, ``{AlpaServe: Statistical Multiplexing with Model Parallelism for Deep Learning Serving},'' in \emph{OSDI}, 2023.

\bibitem{pegasus}
D.~Lo, L.~Cheng, R.~Govindaraju, L.~A. Barroso, and C.~Kozyrakis, ``{Towards energy proportionality for large-scale latency-critical workloads},'' in \emph{ISCA}, 2014.

\bibitem{energyCostDataCenters}
E.~R. Masanet, A.~Shehabi, N.~Lei, S.~J. Smith, and J.~G. Koomey, ``{Recalibrating global data center energy-use estimates},'' \emph{Science}, 2020.

\bibitem{miao2024specinfer}
X.~Miao, G.~Oliaro, Z.~Zhang, X.~Cheng, Z.~Wang, Z.~Zhang, R.~Y.~Y. Wong, A.~Zhu, L.~Yang, X.~Shi, C.~Shi, Z.~Chen, D.~Arfeen, R.~Abhyankar, and Z.~Jia, ``{SpecInfer: Accelerating Generative Large Language Model Serving with Tree-based Speculative Inference and Verification},'' 2024.

\bibitem{twig}
R.~Nishtala, V.~Petrucci, P.~Carpenter, and M.~Sjalander, ``{Twig: Multi-Agent Task Management for Colocated Latency-Critical Cloud Services},'' in \emph{HPCA}, 2020.

\bibitem{dgxh100}
\BIBentryALTinterwordspacing
NVIDIA. {NVIDIA DGX H100}. [Online]. Available: \url{https://www.nvidia.com/en-us/data-center/dgx-h100/}
\BIBentrySTDinterwordspacing

\bibitem{splitwise}
P.~Patel, E.~Choukse, C.~Zhang, A.~Shah, Íñigo Goiri, S.~Maleki, and R.~Bianchini, ``{Splitwise: Efficient generative LLM inference using phase splitting},'' 2023.

\bibitem{polca}
P.~Patel, E.~Choukse, C.~Zhang, Íñigo Goiri, B.~Warrier, N.~Mahalingam, and R.~Bianchini, ``{POLCA: Power Oversubscription in LLM Cloud Providers},'' 2023.

\bibitem{gpuknobscal}
P.~Patel, Z.~Gong, S.~Rizvi, E.~Choukse, P.~Misra, T.~Anderson, and A.~Sriraman, ``Towards improved power management in cloud gpus,'' \emph{IEEE Computer Architecture Letters}, 2023.

\bibitem{wordstowatts}
S.~Samsi, D.~Zhao, J.~McDonald, B.~Li, A.~Michaleas, M.~Jones, W.~Bergeron, J.~Kepner, D.~Tiwari, and V.~Gadepally, ``From words to watts: Benchmarking the energy costs of large language model inference,'' 2023.

\bibitem{scao2022bloom}
T.~L. Scao, A.~Fan, C.~Akiki, E.~Pavlick, S.~Ili{\'c}, D.~Hesslow, R.~Castagn{\'e}, A.~S. Luccioni, F.~Yvon, M.~Gall{\'e} \emph{et~al.}, ``{BLOOM: A 176B-parameter open-access multilingual language model},'' \emph{arXiv preprint arXiv:2211.05100}, 2022.

\bibitem{decoderOnly}
P.~Schmid, ``{Llama 2 is here - Get it on Hugging Face},'' https://huggingface.co/blog/llama2, 2023.

\bibitem{encodeDecode}
P.~Schmid, O.~Sanseviero, P.~Cuence, and L.~Tunstall, ``{Fine-tune FLAN-T5 XL/XXL using DeepSpeed and Hugging Face Transformers},'' https://www.philschmid.de/ fine-tune-flan-t5-deepspeed, 2023.

\bibitem{llama2}
H.~Touvron, L.~Martin, K.~Stone, P.~Albert, A.~Almahairi, Y.~Babaei, N.~Bashlykov, S.~Batra, P.~Bhargava, S.~Bhosale \emph{et~al.}, ``{Llama 2: Open foundation and fine-tuned chat models},'' \emph{arXiv preprint arXiv:2307.09288}, 2023.

\bibitem{transformer}
A.~Vaswani, N.~Shazeer, N.~Parmar, J.~Uszkoreit, L.~Jones, A.~N. Gomez, L.~u. Kaiser, and I.~Polosukhin, ``Attention is all you need,'' in \emph{Advances in Neural Information Processing Systems}, I.~Guyon, U.~V. Luxburg, S.~Bengio, H.~Wallach, R.~Fergus, S.~Vishwanathan, and R.~Garnett, Eds., vol.~30.\hskip 1em plus 0.5em minus 0.4em\relax Curran Associates, Inc., 2017.

\bibitem{orca}
G.-I. Yu, J.~S. Jeong, G.-W. Kim, S.~Kim, and B.-G. Chun, ``{Orca: A Distributed Serving System for Transformer-Based Generative Models},'' in \emph{OSDI}, 2022.

\bibitem{charHW}
H.~Zhang, A.~Ning, R.~Prabhakar, and D.~Wentzlaff, ``{A Hardware Evaluation Framework for Large Language Model Inference},'' 2023.

\bibitem{gemini}
L.~Zhou, L.~N. Bhuyan, and K.~K. Ramakrishnan, ``{Gemini: Learning to Manage CPU Power for Latency-Critical Search Engines},'' in \emph{MICRO}, 2020.

\end{thebibliography}

\end{document}